\documentclass[10pt,twocolumn,letterpaper]{article}

\usepackage{cvpr}              % To produce the CAMERA-READY version
\usepackage[table]{xcolor}
\usepackage{multirow}
\usepackage{supertabular}
\usepackage{lipsum}  % For generating dummy text
\usepackage{afterpage}
\usepackage{float}  % For precise float positioning
\usepackage{dsfont}
\usepackage{bm}
\usepackage{amsmath}
\usepackage{amssymb}
\newcommand{\pub}[1]{\color{black}{\scriptsize{{#1}}}}
\usepackage{multicol}
\usepackage{multirow}
\usepackage{stfloats}
\usepackage{longtable}
\DeclareMathOperator{\sign}{sign}
\usepackage[accsupp]{axessibility}
\usepackage{graphicx}%
\usepackage{amsmath,amssymb,amsfonts}%
\usepackage{amsthm}%
\usepackage{mathrsfs}%
\usepackage[title]{appendix}%
\usepackage{xcolor}%
\usepackage{textcomp}%
\usepackage{manyfoot}%
\usepackage{booktabs}%
\usepackage{algorithm}%
\usepackage{algorithmicx}%
\usepackage{algpseudocode}%
\usepackage{listings}%
\usepackage{bm}
\usepackage[figuresright]{rotating}
\usepackage[justification=centering]{caption}
\theoremstyle{thmstyleone}%
\usepackage{mathtools,array,dcolumn,longtable}
\definecolor{cvprblue}{rgb}{0.21,0.49,0.74}
\usepackage[pagebackref,breaklinks,colorlinks,citecolor=cvprblue]{hyperref}
\newcommand{\M}{DiM}
\title{DiM: $f$-Divergence Minimization Guided Sharpness-Aware Optimization for Semi-supervised Medical Image Segmentation}

\author{
Bingli Wang\\
 Tsinghua University(SZ)\\
\and
Houcheng Su \\
University of Macau\\
\and
Nan Yin \\
Zayed University of Artificial Intelligence\\
United Arab Emirates\\
 yinnan8911@gmail.com
\and
Mengzhu Wang \thanks{Corresponding Author}\\
Hebei University of Technology\\
dreamkily@gmail.com
\and
Li Shen \footnotemark[1]\\
Sun Yat-sen University\\
mathshenli@gmail.com
}
\begin{document}
\maketitle
\begin{abstract}

As a technique to alleviate the pressure of data annotation, semi-supervised learning (SSL) has attracted widespread attention. In the specific domain of medical image segmentation, semi-supervised methods (SSMIS) have become a research hotspot due to their ability to reduce the need for large amounts of precisely annotated data. SSMIS focuses on enhancing the model's generalization performance by leveraging a small number of labeled samples and a large number of unlabeled samples. The latest sharpness-aware optimization (SAM) technique, which optimizes the model by reducing the sharpness of the loss function, has shown significant success in SSMIS. However, SAM and its variants may not fully account for the distribution differences between different datasets. To address this issue, we propose a sharpness-aware optimization method based on $f$-divergence minimization (DiM) for semi-supervised medical image segmentation. This method enhances the model's stability by fine-tuning the sensitivity of model parameters and improves the model's adaptability to different datasets through the introduction of $f$-divergence. By reducing $f$-divergence, the DiM method not only improves the performance balance between the source and target datasets but also prevents performance degradation due to overfitting on the source dataset. 
% Compared to existing state-of-the-art techniques, our method has made groundbreaking progress in the Dice score on the prostate dataset, and this success has been confirmed across three publicly available datasets.
\end{abstract}

\section{Introduction}
\label{sec:intro}

Medical Image Segmentation (MIS)\cite{liu2021feddg, wang2022semi, bortsova2019semi} plays a crucial role in assisting computers with disease diagnosis and treatment research by helping identify key organs or lesions in abnormal images. Recently, numerous supervised learning-based encoder-decoder network architectures have made significant advancements in medical image segmentation, such as U-Net\cite{ronneberger2015u}, U-Net++\cite{zhou2019unet++}, and H-DenseUNet\cite{li2018h}. However, the success of these technologies largely relies on large-scale, pixel-level annotated data. In practice, annotating medical images is not only costly but also challenging due to issues such as low contrast and noise, making it difficult to clearly display images. Moreover, medical images require more specialized knowledge compared to natural images, which makes constructing a large-scale, accurately annotated medical image database nearly an impossible task. In contrast, semi-supervised learning~\cite{miao2023caussl, wang2019semi} offers a new solution to the problem of insufficient data supervision in weakly supervised learning~\cite{zhou2018brief}. It primarily utilizes a small amount of labeled data and a large amount of unlabeled data for joint training. Clearly, semi-supervised learning is significantly more suitable for medical image segmentation and adapting to real-world clinical scenarios than traditional supervised learning methods.

Due to the easy availability of unlabeled data, doctors may not have the time to verify its distribution when faced with massive amounts of data. This "domain shift"~\cite{dash2019big,rudrapatna2020opportunities} issue can lead to significant performance degradation in models, and it is a critical concern when developing semi-supervised medical image segmentation (SSMIS)~\cite{10657413} models. In fact, we should allow unlabeled data to come from one or more different distributions. However, existing unsupervised domain adaptation (UDA)~\cite{zhang2020collaborative,zhao2022uda,guan2021domain} methods do not directly address this issue because they rely on large amounts of labeled source domain data, which is exactly what SSMIS aims to resolve.

Recent studies, such as Sharpness-Aware Minimization (SAM)\cite{foret2020sharpness}, enhance model generalization performance by reducing the sharpness of the loss function. Here, $\mathcal{L}$ represents the loss function to be minimized, and $\theta$ represents the parameters of the neural network. SAM first computes a weight perturbation $\epsilon$ that maximizes the empirical risk $\mathcal{L}(\theta)$, and then minimizes the loss of the perturbed network. In short, SAM aims to reduce the maximum loss near the model parameters $\theta$. Due to the complexity of this minimization-maximization optimization problem, SAM approximates $\mathcal{L}$ with a surrogate loss function $\mathcal{L}_p(\theta)$ for minimization. However, it is important to note that minimizing $\mathcal{L}_p(\theta)$ does not guarantee reaching the flat minimum region for SSMIS~\cite{zhuang2022surrogate}. KL divergence~\cite{van2014renyi} has demonstrated strong performance in SSMIS. The application of KL divergence in SSMIS primarily improves model training efficiency and accuracy by measuring the differences between different probability distributions. This is particularly useful when dealing with limited labeled data and a large amount of unlabeled data, as it helps guide the model to learn more useful information in a semi-supervised setting. For example, MMLBF~\cite{cheng2018image} propose a region-based multi-phase level set method based on KL divergence. Lu et al.\cite{lu2023uncertainty} estimate uncertainty by calculating the Kullback-Leibler divergence between the predictions of the student and teacher models, and directly use this uncertainty to correct the learning of noisy pseudo-labels, rather than setting a fixed threshold to filter pseudo-labels. SwinMM\cite{wang2023swinmm} includes a masked multi-view encoder and a novel proxy task based on mutual learning, which contributes to effective self-supervised pretraining.

However, all of these methods are considered from the perspective of KL divergence, which is highly sensitive to probability values close to zero in the target distribution, often leading to an infinite divergence. In contrast, $f$-divergence can reduce this sensitivity by selecting an appropriate function, making it more stable and robust, especially when dealing with sparse or extreme distributions. In SSMIS, SAM emphasizes achieving stability by controlling the sensitivity of model parameters, while the introduction of f-divergence helps further regulate the model's adaptability across different domains. By minimizing f-divergence, SAM can enhance the balanced performance of the model across both the source and target domains, while avoiding performance degradation due to overfitting the source domain. 
In this work, to overcome the limitation of SAM and explore the full potential of $f$-divergence, we present a novel method $f$-divergence minimization guided sharpness-aware optimization for semi-supervised medical image segmentation (DiM). By consider the $f$-divergence and sharpness-aware minimization, 
which can still be effectively computed even when the support sets of the distributions are different. Our main contributions can be summarized as follows:
\begin{itemize}
\item we analyze the limitations of SAM-like methods and propose $f$-divergence to ensure the model convergence to a flat region with a small loss. 
    \item To the best of our knowledge, this is the first work to apply $f$-divergence constraints to SAM paradigm.
    \item we demonstrate the superior performance
of DiM to state-of-the-arts on three SSMIS benchmarks.
\end{itemize}

\section{Related Work}
\subsection{Semi-supervised Medical Image Segmentation}
% Medical image complexity makes expert annotation both time-consuming and costly \cite{zhuang2013challenges}. Semi-supervised methods aim to address this by leveraging limited labeled data effectively \cite{bortsova2019semi}. 
% Luo et al. \cite{luo2021semi} introduced a dual-task framework that jointly predicts per-pixel segmentation maps and level set representations, incorporating dual-task consistency for enhanced performance. Wu et al. \cite{wu2022mutual} proposed MC-Net++, which utilizes a shared encoder with multiple decoders and a mutual consistency constraint to identify uncertain regions in unlabeled data. Luo et al. \cite{luo2022semi} combined CNNs and Transformers in a cross-teaching scheme, creating an efficient semi-supervised framework. Miao et al. \cite{miao2023caussl} highlighted the importance of model independence between network branches in SSMS.
% Ma et al. \cite{ma2024constructing} addressed performance degradation due to domain shifts in labeled data, proposing Mixed-domain Semi-supervised Medical Image Segmentation (MIDSS) to generate reliable pseudo-labels. 
% Nevertheless, due to the inherent complexity of medical images, these models often exhibit limited generalization ability and convergence instability, which remains a challenging problem.

Due to the complexity of medical images, extensive manual annotation by experts is both challenging and costly \cite{zhuang2013challenges}. To address this, semi-supervised medical image segmentation approaches have emerged as effective solutions that leverage limited labeled data \cite{bortsova2019semi}. Luo et al. \cite{luo2021semi} proposed a dual-task consistency-based semi-supervised framework to simultaneously predict per-pixel segmentation maps and geometrically-aware level set representations, introducing a dual-task consistency regularization to enhance performance. Wu et al. \cite{wu2022mutual} presented MC-Net++, which employs a shared encoder and multiple distinct decoders and introduced a new mutual consistency constraint. This approach statistically identifies uncertain regions, particularly hard regions within unlabeled data. Luo et al. \cite{luo2022semi} incorporated a cross-teaching approach between CNNs and Transformers, resulting in a simple yet efficient semi-supervised learning framework. Miao et al. \cite{miao2023caussl} highlighted the importance of model independence between networks or branches in semi-supervised medical segmentation (SSMS). Ma et al. \cite{ma2024constructing} identified the issue of performance degradation in semi-supervised medical image segmentation due to shared domain distributions, proposing Mixed-domain Semi-supervised Medical Image Segmentation (MIDSS). They emphasized that generating reliable pseudo-labels for unlabeled data is crucial in domain shifts in labeled data. Nevertheless, due to the inherent complexity of medical images, these models often exhibit limited generalization ability and convergence instability, which remains a challenging problem.

\subsection{Unsupervised Domain Adaptation}
Unsupervised Domain Adaptation (UDA) \cite{ganin2015unsupervised,wang2018deep,liu2022deep} aims to adapt models from a labeled source domain to an unlabeled target domain by minimizing the domain shift. This alignment of feature distributions enables knowledge transfer from source to target, enhancing classification performance \cite{ghosn2003bias,weiss2016survey}. Many UDA approaches use a domain classifier to distinguish source from target features, while the feature extractor learns to match feature distributions \cite{long2018conditional,shu2018dirt}. UDA is widely used in tasks like image classification \cite{liu2022deep}, semantic segmentation \cite{sankaranarayanan2018learning}, and object detection \cite{saito2019strong}. Semi-supervised domain adaptation further incorporates a small amount of labeled target data to improve transfer \cite{saito2019semi}.

\subsection{Sharpness-Aware Minimization (SAM) }
Foret et al. \cite{foret2020sharpness} observed that solely minimizing training loss can lead to suboptimal model quality. They proposed Sharpness-Aware Minimization (SAM), which seeks parameters in neighborhoods of uniformly low loss, resulting in a Min-Max optimization problem suitable for gradient descent. Andriushchenko et al. \cite{andriushchenko2022towards} provided theoretical insights on SAM's implicit bias in diagonal linear networks and empirically examined its behavior in nonlinear networks. Zhou et al. \cite{zhou2023imbsam} addressed SAM's limitation in handling class imbalance, particularly overfitting to tail classes, by introducing Imbalanced-SAM (ImbSAM), a class-aware smoothing approach effective in long-tailed classification and semi-supervised anomaly detection tasks. Wang et al. \cite{wang2024u} introduced a model integrating MedSAM with an uncertainty-aware loss function and SharpMin optimizer, enhancing segmentation accuracy and robustness. However, a tailored solution for semi-supervised medical image segmentation remains absent.
\begin{table*}[t]
    \centering
    \renewcommand{\arraystretch}{1.5} % Adds more row spacing
    \setlength{\tabcolsep}{12pt} % Adds more column spacing
    \footnotesize
    \caption{Various commonly used $ f $-divergences with their derivatives and second derivatives.}
    \begin{tabular}{l|c|c|c}
        \hline
        \textbf{$ f $-divergence} & $ f(x) $ & $ f'(x) $ & $ f''(x) $ \\
        \hline \hline
        \textbf{Reverse KL}      & $ x log x $& $ log x + 1 $& $ \frac{1}{x} $ \\
        \hline
        \textbf{Forward KL}      & $ -log x $& $ -\frac{1}{x} $ & $ \frac{1}{x^2} $ \\
        \hline
        \textbf{Jeffrey}         & $ (x - 1) log x $& $ log x + 1 - \frac{1}{x} $& $ \frac{1}{x} + \frac{1}{x^2} $ \\
        \hline
        \textbf{Jensen-Shannon}  & $ -\frac{x + 1}{2} log \frac{x + 1}{2} + \frac{x}{2} log x $& $ \frac{1}{2} log \frac{2x}{x + 1} $& $ \frac{1}{2x(x + 1)} $ \\
        \hline
        \textbf{Pearson}         & $ \frac{(1 - x)^2}{x} $ & $ 1 - \frac{1}{x^2} $ & $ \frac{2}{x^3} $ \\
        \hline
    \end{tabular}
    \label{f-divergence}
\end{table*}

% \subsection{$f$-divergence}
\section{Methods}
\subsection{Aligning Features via $f$-Divergence}
Semi-supervised medical image segmentation is challenging due to the scarcity of labeled data and the high-dimensional complexity of medical images. In this setting, models must leverage both labeled and unlabeled data to learn precise segmentation boundaries. However, limited labeled data can lead to feature drift, where the representations learned from unlabeled data deviate from those based on labeled data. This misalignment between labeled and unlabeled feature distributions reduces segmentation accuracy and limits generalization on unseen data.

To address this, we utilize $ f $-divergence to align the high-dimensional logits from labeled and unlabeled data, constraining the features of unlabeled data based on a limited set of labeled data. Let $ p_{\text{label}} $ and $ p_{\text{unlabel}} $ denote the distributions of logits for the labeled and unlabeled data over a discrete set $ \mathcal{X} $. Our goal is to guide $ p_{\text{unlabel}} $ by minimizing the $ f $-divergence between these distributions in high-dimensional space.

This alignment mitigates feature drift, improving generalization in the target domain. Additionally, $f$-divergence operates effectively in high-dimensional spaces, making it well-suited for capturing subtle but critical variations in medical image features. By aligning feature distributions, $f$-divergence fosters a stable and consistent feature representation, enhancing segmentation accuracy in semi-supervised conditions.

For a convex function $ f(x): \mathbb{R}^{+} \rightarrow \mathbb{R} $ with $ f(1) = 0 $, the $ f $-divergence $ D_f(p_{\text{label}} \| p_{\text{unlabel}}) $ is defined as:
\begin{equation}
    \begin{aligned}
        D_f(p_{\text{label}} \| p_{\text{unlabel}}) &= \mathbb{E}_{x \sim p_{\text{unlabel}}} \left[ f\left( \frac{p_{\text{label}}(x)}{p_{\text{unlabel}}(x)} \right) \right] \\
        &\quad + f'(\infty) p_{\text{label}}(p_{\text{unlabel}} = 0),
    \end{aligned}
\end{equation}
where $ f'(\infty) = \lim_{t \rightarrow 0} t f\left(\frac{1}{t}\right) $. The second term represents the contribution of points $ x $ in the support of $ p_{\text{label}} $ where $ p_{\text{unlabel}}(x) = 0 $, which accounts for cases where the labeled and unlabeled data distributions do not overlap. 

To align $ p_{\text{label}} $ and $ p_{\text{unlabel}} $, we define the alignment loss:
\begin{equation}
\begin{aligned}
\mathcal{L}_{\text{align}} &=D_f(p_{\text{label}} \| p_{\text{unlabel}})  \\
&=\mathbb{E}_{x \sim p_{\text{unlabel}}} \left[ f\left( \frac{p_{\text{label}}(x)}{p_{\text{unlabel}}(x)} \right) \right]
\end{aligned}
\end{equation}

Minimizing $ \mathcal{L}_{\text{align}} $ encourages the logits of unlabeled data to approximate those of labeled data, enhancing feature consistency for semi-supervised learning.

To quantify this alignment, we utilize specific $ f $-divergence variants frequently used in machine learning, including Jeffrey divergence, Jensen-Shannon divergence, and Pearson divergence. Each variant has a unique form of $ f(x) $, $ f'(x) $, and $ f''(x) $, as shown in Table \ref{f-divergence}, enabling flexible divergence calculations between $ p_{\text{label}} $ and $ p_{\text{unlabel}} $. The $ f $-divergences are computed via Monte Carlo estimation based on samples from $ p_{\text{unlabel}} $, applying the respective values in Table \ref{f-divergence} to evaluate $ \mathcal{L}_{\text{align}} $ and facilitate back-propagation during training.

\subsection{Sharpness-Aware Entropy Minimization}
Semi-supervised medical image segmentation leverages both labeled and unlabeled data for accurate boundary detection. However, limited labeled data and distribution shifts often lead to feature inconsistency and unreliable predictions, especially on unseen test samples, necessitating methods that improve model robustness to distributional variations. Sharpness-aware minimization (SAM) enhances model generalization by optimizing within low-loss neighborhoods, stabilizing performance under distribution shifts. However, directly filtering unreliable test samples using gradient norms is challenging due to variations in scale across models and shifts.

Directly using gradient norms to filter out unreliable test samples is challenging due to variability in scale across models and types of distribution shifts. Instead, we leverage entropy as a proxy for gradient magnitude, selecting samples with low entropy values to focus adaptation on confident predictions. Given an entropy function $ E(\bm{x}; \bm{\theta}) $ for a sample $ \bm{x} $ with model parameters $ \bm{\theta} $, we define the selective entropy minimization as:
\begin{equation} 
\label{eq:reliable_entropy}
\begin{aligned}
\min _{\bm{\theta}} S(\bm{x}) E(\bm{x}; \bm{\theta}), \quad
S(\bm{x}) \triangleq \mathbb{I}_{\left\{E(\bm{x}; \bm{\theta}) < E_{0}\right\}}(\bm{x}) 
\end{aligned} 
\end{equation}
where $ S(\bm{x}) $ is an indicator function that activates when the entropy $ E(\bm{x}; \bm{\theta}) $ is below a pre-defined threshold $ E_0 $. This approach ensures that only samples with low entropy (i.e., high confidence) contribute to the training, effectively filtering out unreliable samples that might otherwise induce large gradients.

For further stability, we aim to guide the model towards flatter regions of the entropy loss landscape, which reduces sensitivity to noisy gradients. We define a sharpness-aware entropy objective, $ E^{\text{SA}}(\bm{x}; \bm{\theta}) $, that measures the maximum entropy within a perturbation neighborhood around the current parameters:
% \begin{equation} \label{eq:sa_entropy}
% \begin{aligned}
% &\min _{\bm{\theta}} E^{\text{SA}}(\bm{x}; \bm{\theta}), \quad \\
% &\text{where} \quad E^{\text{SA}}(\bm{x}; \bm{\theta}) \triangleq \max _{\|\bm{\epsilon}\|_{2} \leq \rho} E(\bm{x}; \bm{\theta} + \bm{\epsilon})
% \end{aligned}
% \end{equation}
\begin{equation} 
\label{eq:sa_entropy}
\begin{aligned}
\min _{\bm{\theta}} E^{\text{SA}}(\bm{x}; \bm{\theta}), \quad 
E^{\text{SA}}(\bm{x}; \bm{\theta}) \triangleq \max _{\|\bm{\epsilon}\|_{2} \leq \rho} E(\bm{x}; \bm{\theta} + \bm{\epsilon})
\end{aligned}
\end{equation}
where $ \bm{\epsilon} $ is a perturbation vector constrained within a Euclidean ball of radius $ \rho $. This inner maximization encourages the model to be robust against perturbations, promoting a flat minimum for the entropy loss. Following the SAM approach, we approximate $ \bm{\epsilon}^*(\bm{\theta}) $ by:

\begin{equation} \label{eq:epsilon_approx}
    \hat{\bm{\epsilon}}(\bm{\theta}) = \rho \, \sign\left(\nabla_{\bm{\theta}} E(\bm{x}; \bm{\theta})\right) \frac{|\nabla_{\bm{\theta}} E(\bm{x}; \bm{\theta})|}{\|\nabla_{\bm{\theta}} E(\bm{x}; \bm{\theta})\|_{2}}
\end{equation}

Substituting $ \hat{\bm{\epsilon}}(\bm{\theta}) $ back into the objective, we obtain an approximation for the gradient that encourages flat minima:
\begin{equation} \label{eq:gradient_approx}
    \nabla_{\bm{\theta}} E^{\text{SA}}(\bm{x}; \bm{\theta}) \approx \nabla_{\bm{\theta}} E(\bm{x}; \bm{\theta}) \Big|_{\bm{\theta} + \hat{\bm{\epsilon}}(\bm{\theta})}
\end{equation}

Our final objective for Reliable Sharpness-Aware Entropy Minimization combines selective entropy minimization and sharpness-aware optimization:
\begin{equation} \label{eq:overall_optimization}
    \min _{\tilde{\bm{\theta}}} S(\bm{x}) E^{\text{SA}}(\bm{x}; \bm{\theta})
\end{equation}

In this study, we introduce Sharpness-Aware Entropy Minimization(SAEM), an approach that combines entropy minimization with sharpness-aware training to achieve adaptive entropy reduction, enhancing model stability under challenging conditions. Here, $ S(\bm{x}) $ and $ E^{\text{SA}}(\bm{x}; \bm{\theta}) $ represent entropy measures as defined in Equations \eqref{eq:reliable_entropy} and \eqref{eq:sa_entropy}, respectively. The learnable parameters designated for adaptation are denoted as $ \tilde{\bm{\theta}} \subset \bm{\theta} $.

In essence, SAEM offers a robust framework by integrating entropy filtering with sharpness-aware training, yielding adaptive entropy reduction while ensuring model resilience, particularly under demanding conditions.

\subsection{Loss Function}
The overall loss $\mathcal{L}_{\text{total}}$ is composed of the following components:

1. \textbf{Supervised Loss ($\mathcal{L}_s$)}: Applied to labeled data to guide predictions with ground truth, combining cross-entropy and dice losses for accurate segmentation.

2. \textbf{Intermediate Losses ($\mathcal{L}_{in}$ and $\mathcal{L}_{out}$)}: Defined for intermediate samples $u^{s}_{in}$ and $u^{s}_{out}$, each using weighted cross-entropy ($\mathcal{L}_{ce}$) and dice loss ($\mathcal{L}_{dice}$) to enforce consistency between pseudo labels and model predictions.
\begin{equation}
\begin{aligned}
  \mathcal{L}_{in} = \mathcal{L}_{ce}(\hat{p}_{in}, p^{s}_{in}, w_{in}) + \mathcal{L}_{dice}(\hat{p}_{in}, p^{s}_{in}, w_{in})
\end{aligned}
\end{equation}
    
\begin{equation}
\begin{aligned}
\mathcal{L}_{out} = \mathcal{L}_{ce}(\hat{p}_{out}, p^{s}_{out}, w_{out}) \\
+ \mathcal{L}_{dice}(\hat{p}_{out}, p^{s}_{out}, w_{out})
\end{aligned}
\end{equation}
where $w_{in}$ and $w_{out}$ are pixel-wise weights set by a confidence threshold to filter unreliable pseudo labels.

Each component plays a crucial role in enforcing robust supervision on both labeled and unlabeled data, supporting reliable predictions across domains. The overall loss $\mathcal{L}_{\text{total}}$ is defined as follows:
\begin{equation}
    \mathcal{L}_{\text{total}} = \mathcal{L}_s + \lambda \left( \mathcal{L}_{in} + \mathcal{L}_{out} + \lambda \mathcal{L}_{\text{sym}} \right) + \mathcal{L}_{\text{align}}
\end{equation}
where $\lambda$ is a time-dependent coefficient that scales unsupervised components as training progresses, defined by:
\begin{equation}
    \lambda(t) = e^{-5(1 - t/t_{\text{total}})}.
\end{equation}

\section{Experiments}

\subsection{Experiment Datasets}

\noindent \textbf{Fundus dataset}  consists of retinal fundus images gathered from four medical centers, mainly intended for tasks involving the segmentation of the optic cup and disc. Each image has been cropped to create a region of interest within an 800×800 bounding box. We then resize and randomly crop these images to a size of 256 × 256.

\noindent \textbf{Prostate dataset} includes prostate T2-weighted MRI data, complete with segmentation masks, sourced from six different locations across three public datasets. We randomly divide the dataset into training and testing sets at a ratio of 4:1, resizing and randomly cropping each 2D slice to 384 × 384. Labeled samples are chosen from consecutive slices within individual cases, ensuring there is at most one case overlap and no overlap of slices with unlabeled samples.

\subsection{Comparison Methods and Settings}

Our method is implemented in PyTorch and utilizes an NVIDIA GeForce RTX 3090 GPU. We establish default experimental parameters for training. Optimization is performed using the SAM optimizer, with a base optimizer of Stochastic Gradient Descent (SGD) set to a momentum of 0.9, a weight decay of 0.0001, and an initial learning rate of 0.03. The batch size is set to 8, comprising 4 labeled and 4 unlabeled samples. We conduct a total of 30,000 iterations for the Fundus dataset and 60,000 iterations for the Prostate dataset.

During testing, the final segmentation results are generated by the student model. Our approach is benchmarked against several state-of-the-art (SOTA) methods, including supervised techniques such as UA-MT~\cite{yu2019uncertainty}, FixMatch~\cite{sohn2020fixmatch}, CPS~\cite{chen2021semi}, CoraNet~\cite{shi2021inconsistency}, SS-Net~\cite{wu2022exploring}, BCP~\cite{bai2023bidirectional}, CauSSL~\cite{miao2023caussl}, and MiDSS~\cite{10657413}, as well as domain-unsupervised adaptation methods like FDA~\cite{yang2020fda}, SIFA~\cite{chen2020unsupervised}, and UDA-VAE++~\cite{lu2022unsupervised}.

In each experiment, a limited amount of data from a designated domain (e.g., Domain 1 in \cref{fundus}) is labeled, while the remaining data are treated as unlabeled. For the upper-bound comparison, we utilized the f-divergence, specifically employing the Jensen-Shannon divergence to calculate the distance between logits, and used the most naive SAM optimizer in our experiments. For the upper bound, we followed the results of the MiDSS paper, which applied UCP within the FixMatch framework, utilizing all available training data from a specific domain as labeled data, providing the model with comprehensive source domain information.

Our evaluation metrics include the Dice coefficient (DC), Jaccard coefficient (JC), 95\% Hausdorff Distance (HD), and Average Surface Distance (ASD). Except for SIFA, which incorporates ResNet blocks~\cite{he2016deep} for its generator and decoder, all methods employ the U-Net backbone~\cite{ronneberger2015u}.

\subsection{Comparison with State-of-the-Art Methods}
\noindent  \textbf{Results on Fundus dataset.} With only 20 labeled samples, \M~achieves superior performance across all domains in the optic cup/disc segmentation task, as illustrated in Table~\ref{fundus}. \M~ consistently outperforms competing methods in all metrics, which achieves the highest average DC and JC scores while maintaining the lowest HD and ASD, highlighting its robustness and precision in segmenting dual objects with overlapping regions. These results suggest that \M~effectively mitigates issues faced by other semi-supervised and unsupervised domain adaptation methods, such as error accumulation and limited knowledge transfer, ensuring both high accuracy and generalizability across multiple domains.

\noindent  \textbf{Results on Prostate dataset.} As shown in Table~\ref{prostate}, \M~achieves outstanding performance across all metrics on Prostate dataset. It is also noteworthy that \M~achieves the highest DC and JC averages while maintaining the lowest HD and ASD scores, suggesting higher segmentation accuracy and boundary precision. The inclusion of SAM likely contributes to improved generalization by mitigating sharp minima, while f-divergence loss enhances alignment of the predicted and true distributions, reducing segmentation errors. These results underscore the robustness and effectiveness of our method.

\begin{table*}[h]
\centering

\footnotesize
\resizebox{\textwidth}{!}{
\begin{tabular}{ll|c|cccc|cccc}
    \hline
    \multicolumn{3}{c|}{Task} & \multicolumn{8}{c}{Optic Cup / Disc Segmentation}\\
    \hline
    \multicolumn{2}{c|}{\multirow{2}{*}{Method}} & \multirow{2}{*}{\#L} & \multicolumn{4}{c|}{DC $\uparrow$} & DC $\uparrow$ & JC $\uparrow$ & HD $\downarrow$ & ASD $\downarrow$\\
    \cline{4-11}
    & & & Domain 1 & Domain 2 & Domain 3 & Domain 4 & Avg. & Avg. & Avg. & Avg.\\
    \hline
    U-Net & & 20 & 59.54 / 73.89 & 71.28 / 74.23 & 50.87 / 64.29 & 35.61 / 63.30 & 61.63 & 52.65 & 48.28 & 28.86\\
    UA-MT & \pub{MICCAI'19} & 20 & 59.35 / 78.46 & 63.08 / 74.45 & 35.24 / 47.73 & 36.18 / 55.43 & 56.24 & 47.00 & 48.64 & 31.35\\
    FDA & \pub{CVPR'20} & 20 & 76.99 / 89.94 & \underline{77.69} / 89.63 & 78.27 / 90.96 & 64.52 / 74.29 & 80.29 & 71.05 & 16.23 & 8.44\\
    SIFA & \pub{TMI'20} & 20 &50.67 / 75.30 & 64.44 / 80.69 & 61.67 / 83.77 & 55.07 / 70.67 & 67.78 & 54.77 & 20.16 & 10.93\\
    FixMatch & \pub{NeurIPS'20} & 20 & 81.18 / 91.29 & 72.04 / 87.60 & 80.41 / 92.95 & 74.58 / 87.07 & 83.39 & 73.48 & 11.77 & 5.60\\
    CPS & \pub{CVPR'21} & 20 & 64.53 / 86.25 & 70.26 / 86.97 & 42.92 / 54.94 & 36.98 / 46.70 & 61.19 & 52.69 & 34.44 & 26.79\\
    CoraNet & \pub{TMI'21} & 20 & 61.64 / 87.32 & 65.56 / 87.05 & 66.12 / 83.54 & 49.01 / 77.73 & 72.25 & 60.50 & 20.52 & 10.44\\
    UDA-VAE++ & \pub{CVPR'22} & 20 & 55.01 / 80.76 & 68.87 / 85.94 & 63.23 / 84.92 & 68.42 / 80.89 & 73.51 & 61.40 & 17.60 & 9.86\\
    SS-Net & \pub{MICCAI'22} & 20 & 59.42 / 78.15 & 67.32 / 85.05 & 45.69 / 69.91 & 38.76 / 61.13 & 63.18 & 53.49 & 44.90 & 25.73\\
    BCP & \pub{CVPR'23} & 20 & 71.65 / 91.10 & 77.19 / \textbf{92.00} & 72.63 / 90.77 & 77.67 / 91.42 & 83.05 & 73.66 & 11.05 & 5.80\\
    CauSSL & \pub{ICCV'23} & 20 & 63.38 / 80.60 & 67.52 / 80.72 & 49.53 / 63.88 & 39.43 / 49.43 & 61.81 & 51.80 & 41.25 & 23.94\\
    MiDSS & \pub{CVPR'24}& 20 & \underline{83.39} / \textbf{92.96} & 73.12 / 88.88 & \underline{83.50} / \underline{92.97} & \underline{78.63} / \textbf{93.38} & \underline{85.85} & \underline{76.95} & \underline{9.06} & \underline{4.40}\\
    \hline
    \M &  & 20 & \textbf{84.68} / \underline{92.91}& \textbf{78.16} / \underline{90.49}& \textbf{84.82 / 93.36}& \textbf{81.63} / \underline{92.18}& \textbf{87.28}& \textbf{78.53}& \textbf{7.83}& \textbf{3.82}\\
    \color{gray} Upper bound & & \color{gray} * & \color{gray} 85.53 / \color{gray} 93.41 & \color{gray} 80.55 / \color{gray} 90.90 & \color{gray} 85.44 / \color{gray} 93.04 & \color{gray} 85.61 / \color{gray} 93.21 & \color{gray} 88.46 & \color{gray} 80.35 & \color{gray} 7.41 & \color{gray} 3.70\\

    \hline
\end{tabular}
}

\caption{Comparison of methods on the Fundus dataset. \#L indicates the number of labeled samples. In the "Upper bound" row, * denotes using all training samples in a domain as labeled data. An upward arrow (↑) signifies that higher values indicate better performance, while a downward arrow (↓) indicates the opposite. The best results are bolded, with the second-best underlined.}
\label{fundus}

\end{table*}

\begin{table*}[h]
\centering

\footnotesize
\resizebox{\textwidth}{!}{
    \begin{tabular}{ll|c|cccccc|cccc}
    \hline
    \multicolumn{3}{c|}{Task} & \multicolumn{10}{c}{Prostate Segmentation}\\
    \hline
    \multicolumn{2}{c|}{\multirow{2}{*}{Method}} & \multirow{2}{*}{\#L} & \multicolumn{6}{c|}{DC $\uparrow$} & DC $\uparrow$ & JC $\uparrow$ & HD $\downarrow$ & ASD $\downarrow$\\
    \cline{4-13}
    & & & RUNMC & BMC & HCRUDB & UCL & BIDMC & HK & Avg. & Avg. & Avg. & Avg.\\
    \hline
    U-Net & & 40 & 31.11 & 35.07 & 20.04 & 38.18 & 19.41 & 26.62 & 28.41 & 23.24 & 95.11 & 65.84\\
    UA-MT & \pub{MICCAI'19} & 40 & 29.44 & 4.68 & 12.49 & 39.42 & 17.94 & 18.22 & 20.37 & 14.88 & 112.07 & 77.58\\
    FDA & \pub{CVPR'20} & 40 & 47.44 & 35.37 & 24.54 & 61.01 & 28.19 & 40.51 & 39.51 & 32.17 & 76.67 & 47.87\\
    SIFA & \pub{TMI'20} & 40 & 72.67 & 70.37 & 64.08 & 73.49 & 71.62 & 65.16 & 69.57 & 56.78 & 29.43 & 13.03\\
    FixMatch & \pub{NeurIPS'20} & 40 & 83.58 & 69.17 & 73.63 & 79.21 & 56.07 & 84.78 & 74.41 & 65.96 & 24.18 & 14.09\\
    CPS & \pub{CVPR'21} & 40 & 29.83 & 9.21 & 11.84 & 43.84 & 13.51 & 14.56 & 20.47 & 15.12 & 115.96 & 78.51\\
    CoraNet & \pub{TMI'21} & 40 & 69.43 & 31.16 & 16.29 & 69.33 & 24.66 & 22.16 & 38.84 & 31.48 & 67.91 & 44.98\\
    UDA-VAE++ & \pub{CVPR'22} & 40 & 68.73 & 69.36 & 65.49 & 67.19 & 63.29 & 65.15 & 66.54 & 52.80 & 34.20 & 15.48\\
    SS-Net & \pub{MICCAI'22} & 40 & 29.10 & 13.49 & 14.20 & 51.96 & 23.83 & 13.23 & 24.30 & 18.74 & 109.54 & 71.13\\
    BCP & \pub{CVPR'23} & 40 & 70.15 & 71.97 & 46.15 & 58.93 & 74.21 & 67.47 & 64.81 & 55.17 & 52.60 & 27.22\\
    CauSSL & \pub{ICCV'23} & 40 & 24.10 & 27.46 & 16.94 & 27.23 & 15.28 & 14.56 & 20.93 & 15.48 & 114.62 & 73.30\\
    MiDSS& \pub{CVPR'24}& 40 & \underline{87.94} & \underline{85.30} & \underline{77.74} & \underline{86.29}& \textbf{88.54} & \underline{86.43} & \underline{85.37}& \underline{77.30}& \underline{13.44}& \underline{6.18}\\
    \hline
    \M & & 40 & \textbf{88.43}& \textbf{85.67}& \textbf{87.56}&  \textbf{87.27}&  \underline{88.23}&  \textbf{87.55}& \textbf{87.45}& \textbf{79.52}& \textbf{10.57}& \textbf{4.35}\\
    \color{gray} Upper bound & & \color{gray} * & \color{gray} 88.52 & \color{gray} 88.61 & \color{gray} 85.71 & \color{gray} 88.61 & \color{gray} 88.98 & \color{gray} 89.49 & \color{gray} 88.32 & \color{gray} 80.71 & \color{gray} 10.05 & \color{gray} 4.12\\

    \hline

\end{tabular}
}

\caption{Comparison of different methods on Prostate dataset.}
\label{prostate}
\end{table*}

\subsection{Ablation Study}

Firstly, We conduct ablation studies to show the impact of each component in \M~. Sencondly, our objective is to assess whether varying types of \( f \)-divergences lead to notable performance differences in the model and to identify the optimal form for superior outcomes.

\noindent \textbf{The effectiveness of SAM.} As demonstrated in Table~\ref{fundusablation}, incorporating the Sharpness-Aware Minimization (SAM) optimizer (as seen in Method \#2 and \#4 compared to \#1.) enhances model performance on the Optic Cup/Disc segmentation task. SAM effectively reduces the model’s loss, increasing robustness to minor data distribution shifts and enabling more efficient capture of inter-sample similarity. Consequently, SAM improves the DC and JC scores while reducing the HD and ASD. These results indicate that SAM not only strengthens the model's generalization ability but also enhances segmentation accuracy and boundary precision.

\noindent  \textbf{The effectiveness of $f$-Divergence.} The incorporation of $f$-divergence (as seen in Method \#3 and \#4 compared to \#1.) contributes to notable improvements in model performance, particularly through a marked reduction in ASD and HD metrics, as shown in Table~\ref{fundusablation}. By quantifying the discrepancy between probability distributions of labeled and unlabeled data, $f$-divergence enhances the model's capacity to represent features within the unlabeled dataset. Experimental results demonstrate that introducing $f$-divergence allows the model to more precisely capture the boundaries of structurally similar regions. This results in further gains in DC and JC metrics, along with substantial reductions in HD and ASD, thereby indicating improved accuracy in segmentation, especially along edge details.

\begin{table}[t]
\centering
% \footnotesize
% \setlength{\tabcolsep}{3pt} % 调整列间距
\resizebox{\columnwidth}{!}{
\begin{tabular}{c|ccc|cccc}
    \hline
    \multicolumn{4}{c|}{Task} & \multicolumn{4}{c}{Optic Cup / Disc Segmentation}\\
    \hline
    \multirow{2}{*}{Method} & \multirow{2}{*}{Baseline} & \multirow{2}{*}{SAM} & \multirow{2}{*}{$f$-Divergence}& \multirow{2}{*}{DC $\uparrow$} & \multirow{2}{*}{JC $\uparrow$} & \multirow{2}{*}{HD $\downarrow$} & \multirow{2}{*}{ASD $\downarrow$} \\

    & & & &  &  &  &   \\
    \hline
    \#1 & \checkmark & & & 88.27 & 80.02 & 7.19 & 3.48 \\
    \#2 & \checkmark & \checkmark & & 88.33 & 80.22 & 7.10 & 3.52 \\
    \#3 & \checkmark & & \checkmark & 88.32 & 80.14 & 7.22 & 3.47  \\
    \hline
    \#4 & \checkmark & \checkmark & \checkmark & \textbf{88.80} & \textbf{80.74} & \textbf{6.81} & \textbf{3.27}  \\
    \hline
\end{tabular}
}
\caption{Ablation experiments across domain 1 on Fundus dataset.}
\label{fundusablation}
\end{table}

\noindent  \textbf{$f$-Divergence strategies.} Based on the experimental results shown in the Table~\ref{domain_metrics_consolidated}, different $f$-divergence strategies demonstrate varying degrees of effectiveness for optic cup and disc segmentation across four domains in the Fundus dataset. Generally, the JS divergence and Jeffrey divergence strategies perform well, often yielding higher DC and JC while reducing HD and ASD. Specifically, JS divergence tends to provide more consistent results in edge cases, as evidenced by lower HD and ASD values, while also achieving competitive segmentation accuracy. Pearson divergence, on the other hand, exhibits a balanced performance, particularly excelling as the second-best in some metrics.

\begin{table*}[h]
    \centering
    \begin{tabular}{c|c|cc|cc|cc|cc}
    \hline
    \multicolumn{2}{c|}{Task} & \multicolumn{8}{c}{Optic Cup / Disc Segmentation} \\
    \hline
    \multirow{2}{*}{Type} & \multirow{2}{*}{Domain} & \multicolumn{2}{c|}{DC $\uparrow$} & \multicolumn{2}{c|}{JC $\uparrow$} & \multicolumn{2}{c|}{HD $\downarrow$} & \multicolumn{2}{c}{ASD $\downarrow$} \\
    \cline{3-10}
    & & Cup & Disc & Cup & Disc & Cup & Disc & Cup & Disc \\
    \hline
    % \multirow{1}{*}{3(5\%)}
    baseline & \multirow{4}{*}{Domain 1} & 83.38 & \textbf{93.15} & 72.68 & \textbf{87.36} & 8.28 & \underline{6.10} & 4.01 & \textbf{2.95} \\
    JS &  & \textbf{84.68} & 92.91 & \textbf{74.51} & 86.96 & \textbf{7.63} & \textbf{5.98} & \textbf{3.56} & \underline{2.97}\\
    Jeffrey &   & 82.82 & 92.57 & 72.07 & 86.45 & 8.20 & 6.21 & 4.13 & 3.20 \\
    Pearson &   & \underline{83.54} & \underline{92.94} & \underline{73.15} & \underline{87.07} & \underline{7.91} & 6.23 & \underline{3.94} & 3.09\\ 
    \hline
    baseline & \multirow{4}{*}{Domain 2} & 73.11 & 88.88 & 59.76 & 80.78 & 12.89 & 13.69 & 6.56 & 6.11 \\
    JS &   & 78.10 & 89.32 & \underline{65.54} & 81.28 & 10.71 & 13.72 & 5.29 & 6.21 \\
    Jeffrey &  & \underline{78.16} & \textbf{90.49} & 65.44 & \textbf{82.99} & \textbf{9.62} & \textbf{8.24} & \textbf{4.83} & \textbf{4.37} \\
    Pearson &  & \textbf{78.50} & \underline{90.07} & \textbf{66.29} & \underline{82.41} & \underline{10.62} & \underline{10.67} & \underline{5.19} & \underline{5.11} \\
    \hline
    baseline & \multirow{4}{*}{Domain 3}  & 83.49 & \textbf{93.36} & \underline{72.77} & \textbf{87.82} & \underline{8.06} & \textbf{6.19} & \underline{3.89} & \textbf{3.15} \\
    JS &  & \textbf{84.82} & \textbf{93.36} & \textbf{74.66} & \underline{87.53} & \textbf{7.56} & 6.24 & \textbf{3.62} & \textbf{3.15} \\
    Jeffrey &  & \underline{83.58} & 93.12 & 72.71 & 87.40 & 8.78 & \underline{6.23} & 4.00 & \underline{3.17} \\
    Pearson &  & 83.03 & \underline{93.14} & 72.02 & 87.43 & 9.08 & \underline{6.23}& 4.20 & 3.19 \\
    \hline
    baseline & \multirow{4}{*}{Domain 4} & 78.63& \textbf{93.56} & 66.27& \textbf{88.16} & 10.95& 6.28& 5.44& \textbf{3.06} \\
    JS &  & \textbf{81.63} & 92.18 & \textbf{70.23} & 85.87 & \textbf{9.32} & 8.01& \textbf{4.42} & 3.87 \\
    Jeffrey &  & \underline{79.24}& \underline{93.31}& 66.86& \underline{87.74}& 10.48& \textbf{6.15}& 5.00& \underline{3.16}\\
    Pearson &  & 80.03& 93.12& \underline{67.92}& 87.43& \underline{10.51}& \underline{6.31}& \underline{4.93}& 3.28\\
    \hline
    \end{tabular}
    \caption{Performance comparison of different $f$-divergence strategies across four domains on Fundus dataset. Metrics marked with $\uparrow$ indicate that higher values imply better performance, while those with $\downarrow$ suggest the opposite. The best performance results are highlighted in bold, and the second-best are underlined.}
    \label{domain_metrics_consolidated}
\end{table*}

\subsection{Visualization Analysis}

\begin{figure}[ht]
    \centering

    \subfloat[MiDSS]{\includegraphics[width=0.50\columnwidth]{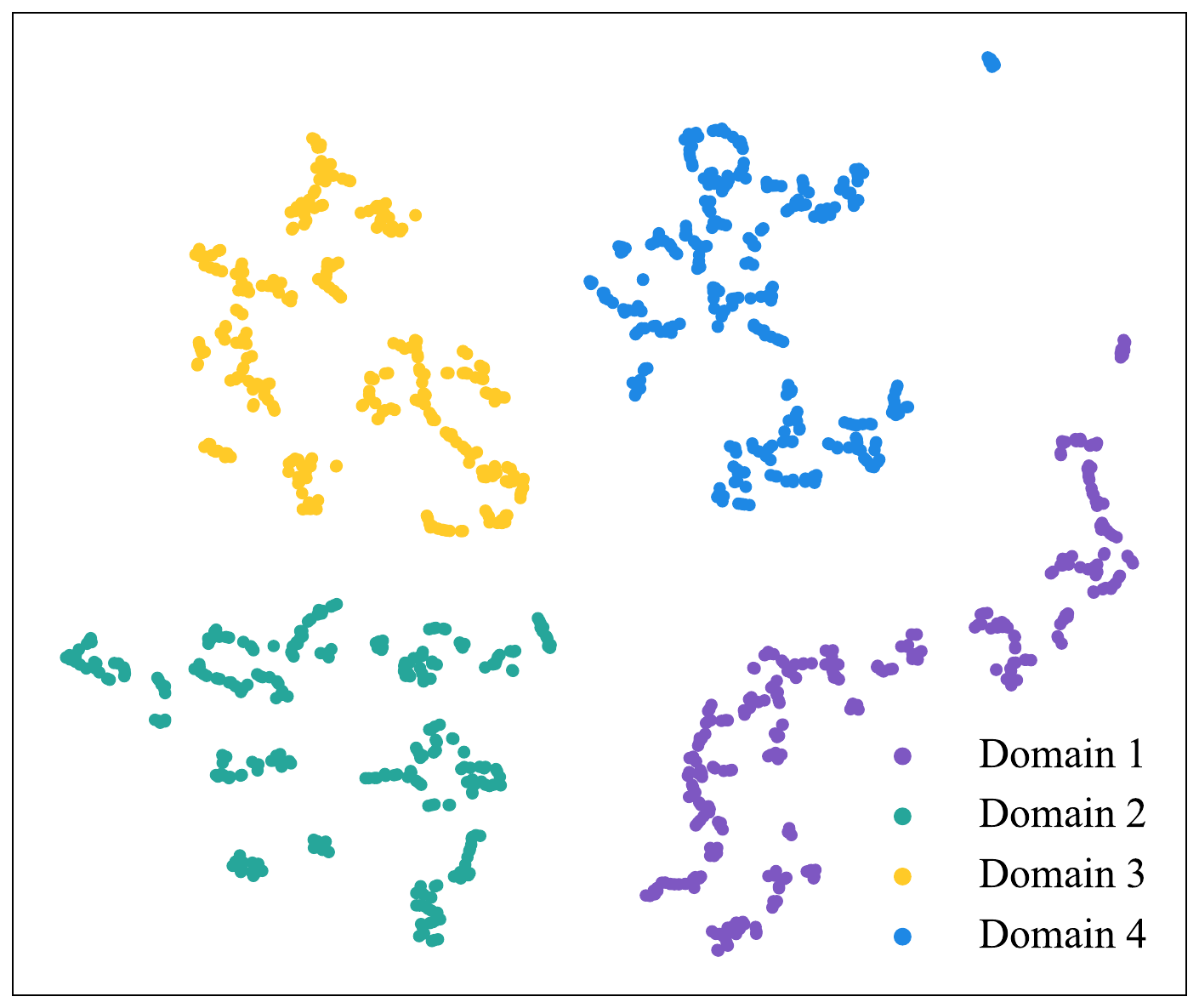}}
    \subfloat[~\M]{\includegraphics[width=0.5\columnwidth]{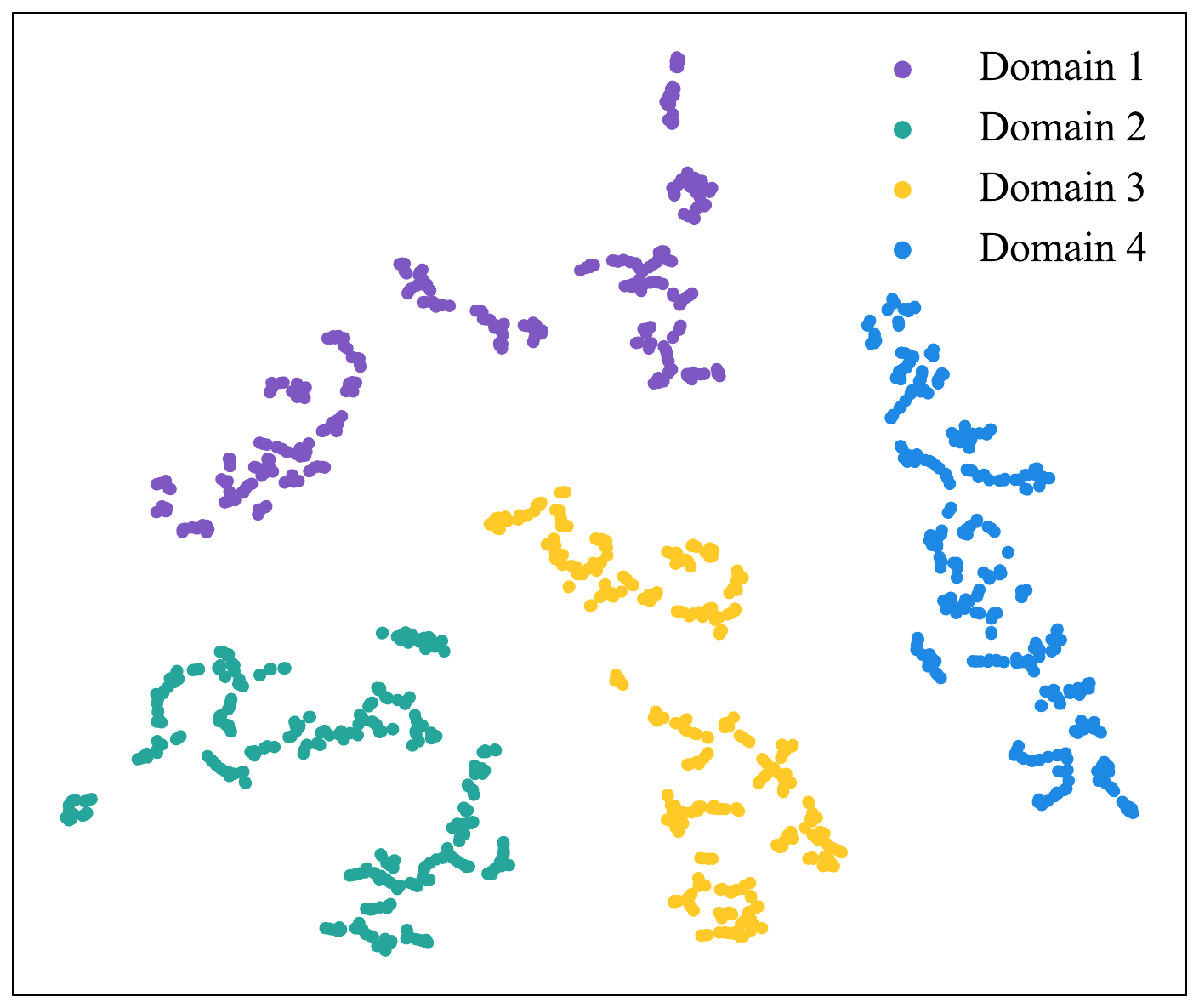}}
    \vspace{-2mm}
    
    \caption{A T-sne visualization analysis was performed on the Fundus dataset experiment.}
    \label{fig:visual1}
     \vspace{-2mm}
\end{figure}

\begin{figure*}
    \centering
    \includegraphics[width=1\linewidth]{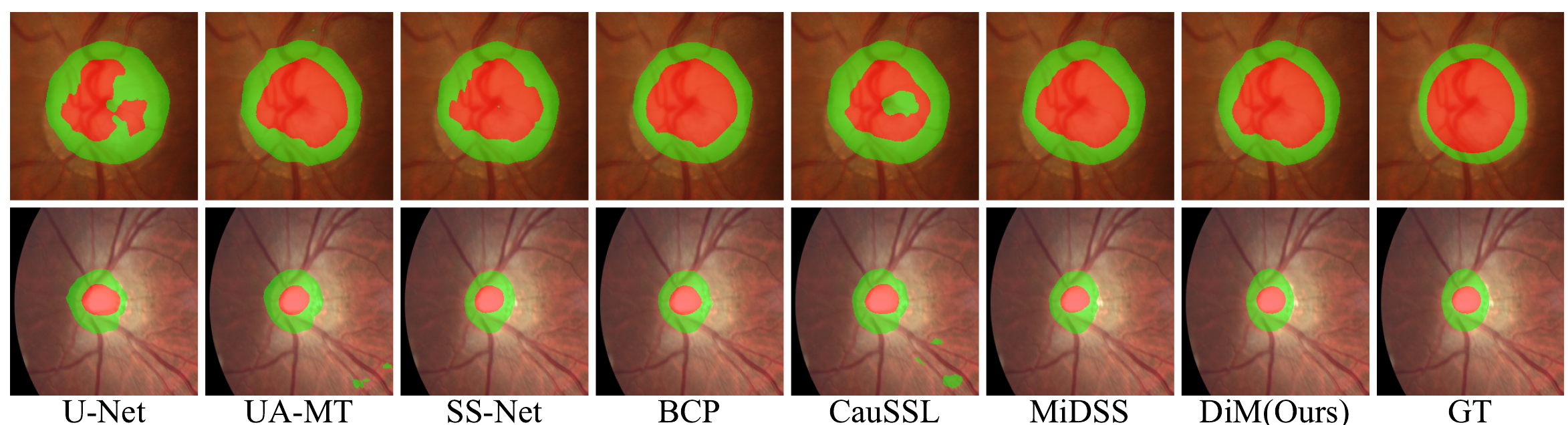}
    \caption{Visual comparison of segmentation results on Fundus dataset across different models. Red and green represent the Optical Cup and Disc, respectively. The first row shows segmentation results on test samples from the labeled domain (Domain 1), while the second row presents results on test samples from a different domain (Domain 4).}
    \label{fig:fundus-visulization}
\end{figure*}

\begin{figure*}
    \centering
    \includegraphics[width=1\linewidth]{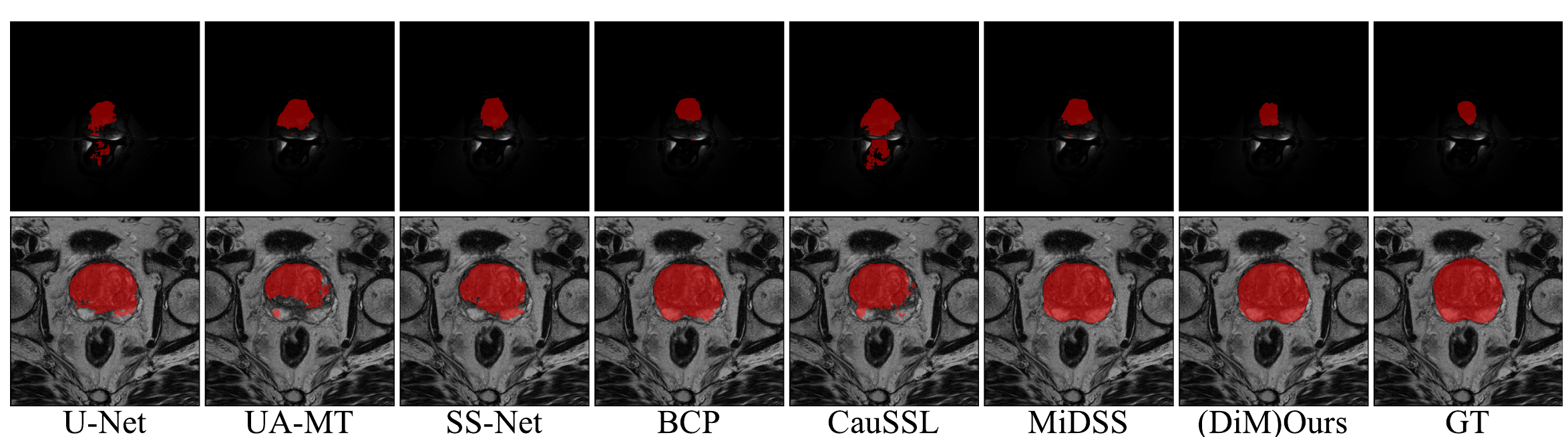}
    \caption{Visual results on Prostate dataset. }
    \label{fig:prostate-visulization}
\end{figure*}
\noindent  \textbf{T-SNE visualization.} We adopt the T-SNE visualization method , which graphically represents the learning representation obtained from our method, as shown in Figure.~\ref{fig:visual1}(a) and ~\ref{fig:visual1}(b). \M~significantly improves the feature alignment between different domains compared to MiDSS, resulting in a tighter and more coherent data distribution. 

\noindent  \textbf{Segmentation visualization.} As shown in Figure~\ref{fig:fundus-visulization} , \M~achieves higher accuracy and better preservation of edge details in optic disc and cup segmentation tasks compared to other methods. Models such as U-Net, UA-MT, and SS-Net show noticeable boundary blurring, with segmentation results that lack precision, particularly at the structural boundaries of the optic disc and cup. These models tend to either miss segments or over-segment certain regions. In contrast, our model produces clearer boundaries with more precise edge detail retention, and the segmentation within the optic disc and cup regions is more cohesive, closely matching the ground truth (GT). 

We also conducted visualization experiments on the prostate segmentation task in Figure~\ref{fig:prostate-visulization}. Results indicate that \M~continues to outperform other methods, such as BCP, CauSSL, and MiDSS, by achieving clearer boundary delineation and better edge detail preservation. The segmentation closely matches the GT, demonstrating improved accuracy and cohesion within the prostate region, even in challenging boundary areas.

\noindent  \textbf{Model Performance.} The validation loss and Dice coefficient curves across the four domains on the Fundus dataset demonstrate stable model performance, which are depicted in Figure~\ref{fig:loss_dice}. Loss decreases rapidly in the early epochs and converges across all domains, indicating effective training. Dice scores for both the optic cup and disc steadily increase and plateau at high values, with the optic disc achieving near-perfect accuracy. 

% \begin{figure}
%     \centering
%     \includegraphics[width=1\linewidth]{domains_loss_dice.pdf}
%     \caption{Enter Caption}
%     \label{fig:enter-label}
% \end{figure}

% \begin{figure*}[ht]
%     \centering
    
%     % 第一行
%     \subfloat[Domain 1 Loss]{\includegraphics[width=0.5\columnwidth]{domain_1_loss.pdf}}
%     \subfloat[Domain 1 Dice]{\includegraphics[width=0.5\columnwidth]{domain_1_dice.pdf}} \\
    
%     % 第二行
%     \subfloat[Domain 2 Loss]{\includegraphics[width=0.5\columnwidth]{domain_2_loss.pdf}}
%     \subfloat[Domain 2 Dice]{\includegraphics[width=0.5\columnwidth]{domain_2_dice.pdf}} \\
    
%     % 第三行
%     \subfloat[Domain 3 Loss]{\includegraphics[width=0.5\columnwidth]{domain_3_loss.pdf}}
%     \subfloat[Domain 3 Dice]{\includegraphics[width=0.5\columnwidth]{domain_3_dice.pdf}} \\
    
%     % 第四行
%     \subfloat[Domain 4 Loss]{\includegraphics[width=0.5\columnwidth]{domain_4_loss.pdf}}
%     \subfloat[Domain 4 Dice]{\includegraphics[width=0.5\columnwidth]{domain_4_dice.pdf}}
    
%     \vspace{-2mm}
%     \caption{Validation Loss and Dice across Domains on the Fundus}
%     \label{fig:visual1}
%     \vspace{-2mm}
% \end{figure*}

\begin{figure*}[ht]
    \centering
    
    % 第一行
    \subfloat[Domain 1 Loss]{\includegraphics[width=0.5\columnwidth]{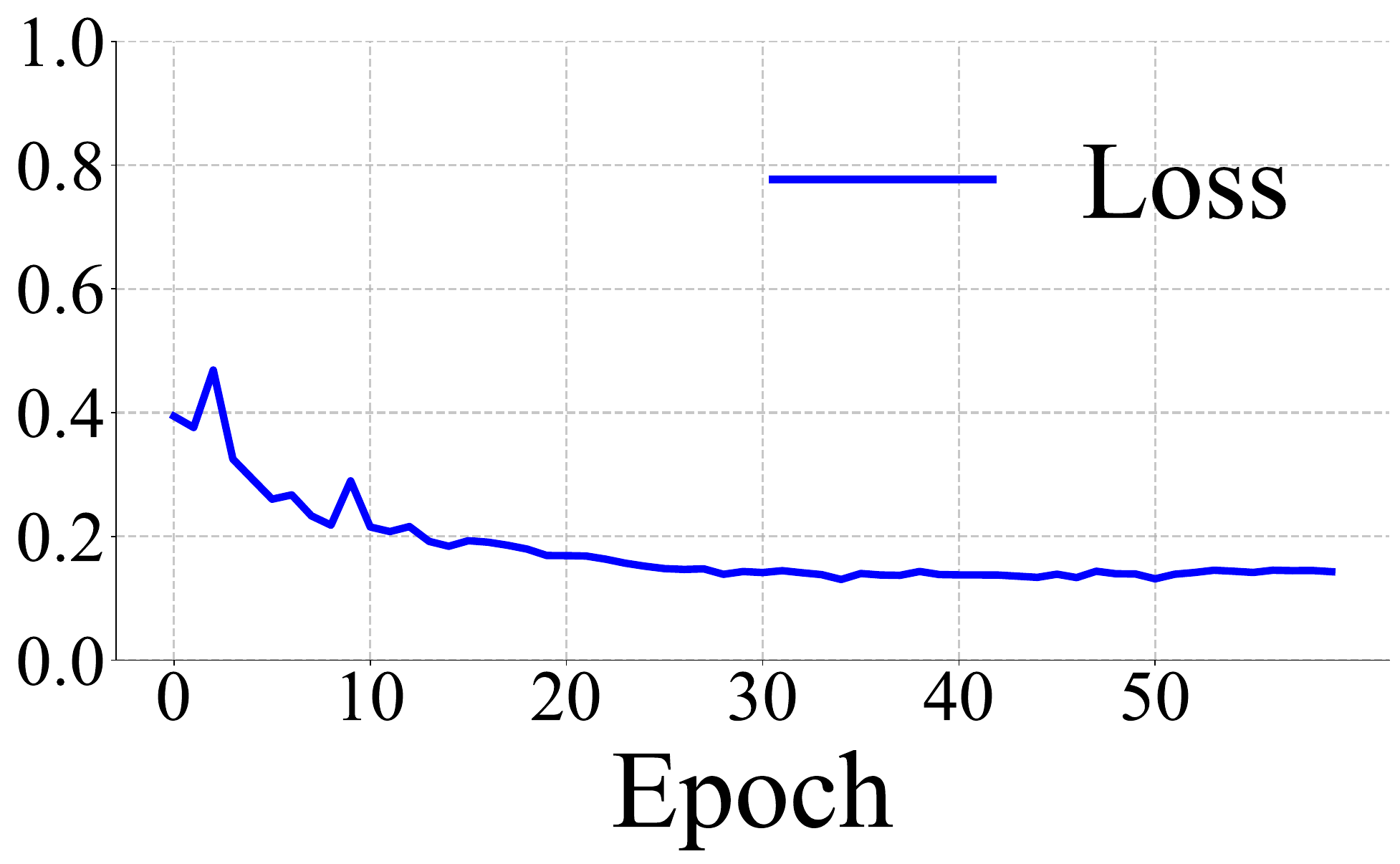}}
    \subfloat[Domain 2 Loss]{\includegraphics[width=0.5\columnwidth]{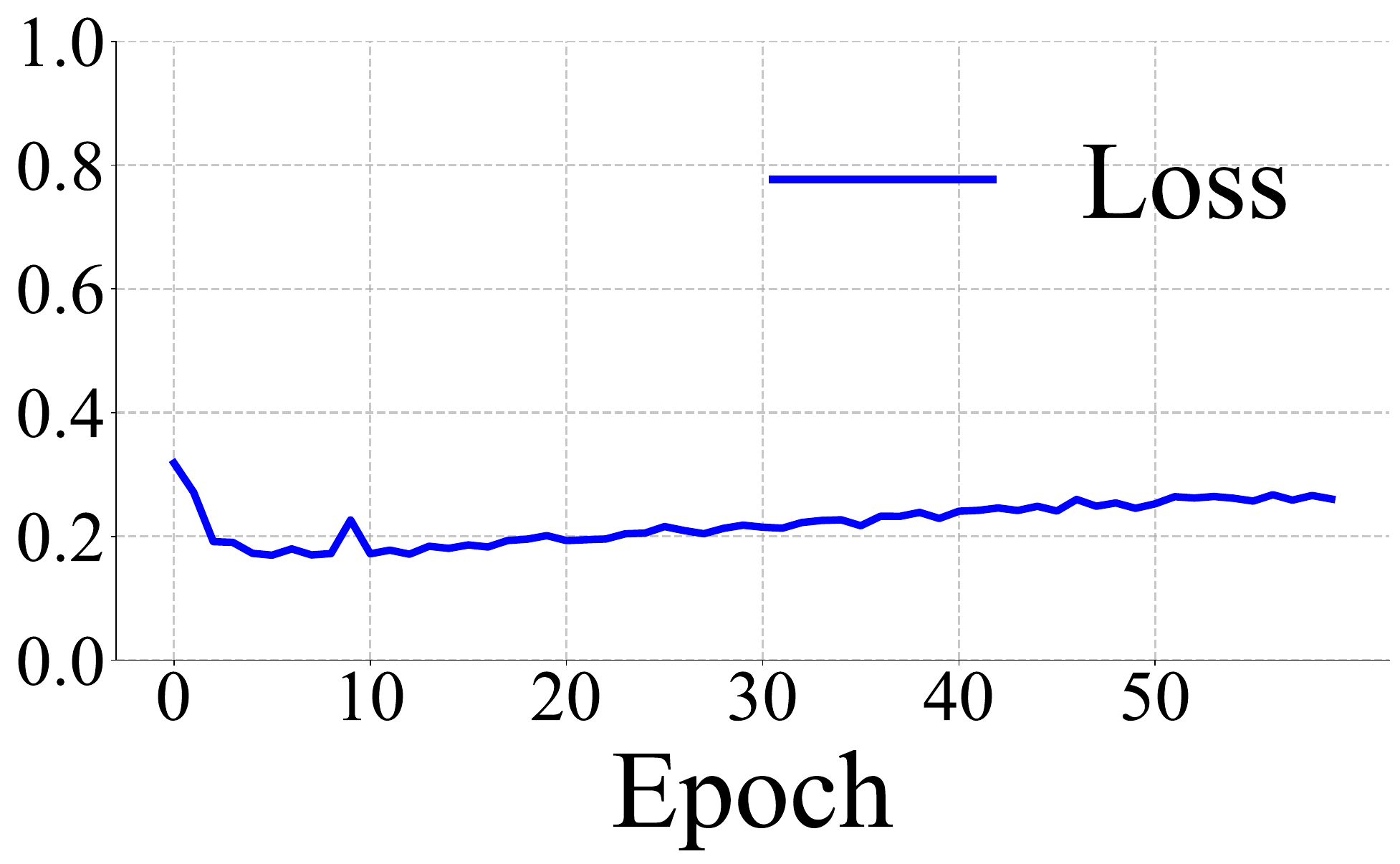}}
    \subfloat[Domain 3 Loss]{\includegraphics[width=0.5\columnwidth]{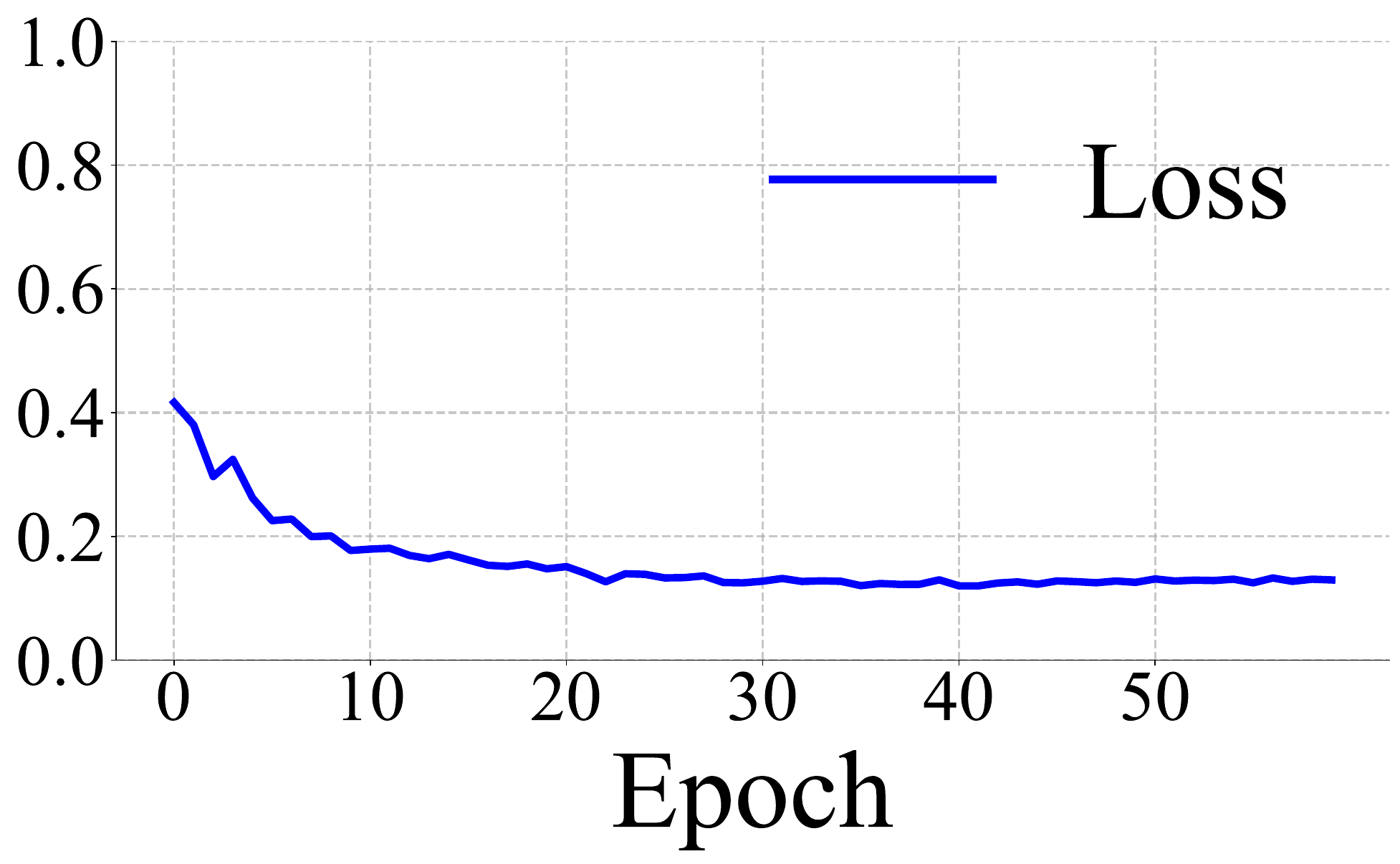}}
    \subfloat[Domain 4 Loss]{\includegraphics[width=0.5\columnwidth]{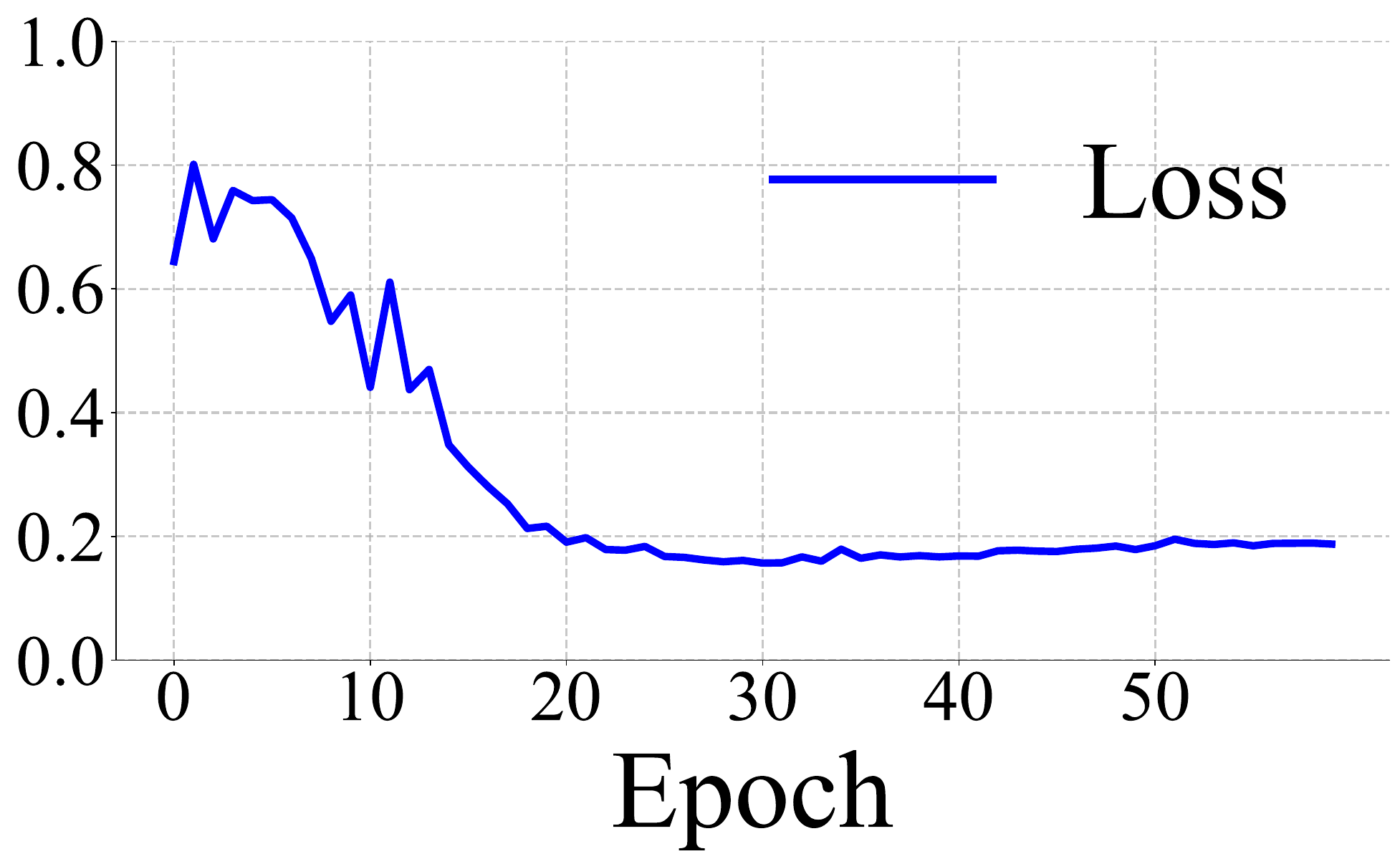}}
     \\
    
    % 第二行
    \subfloat[Domain 1 Dice]{\includegraphics[width=0.5\columnwidth]{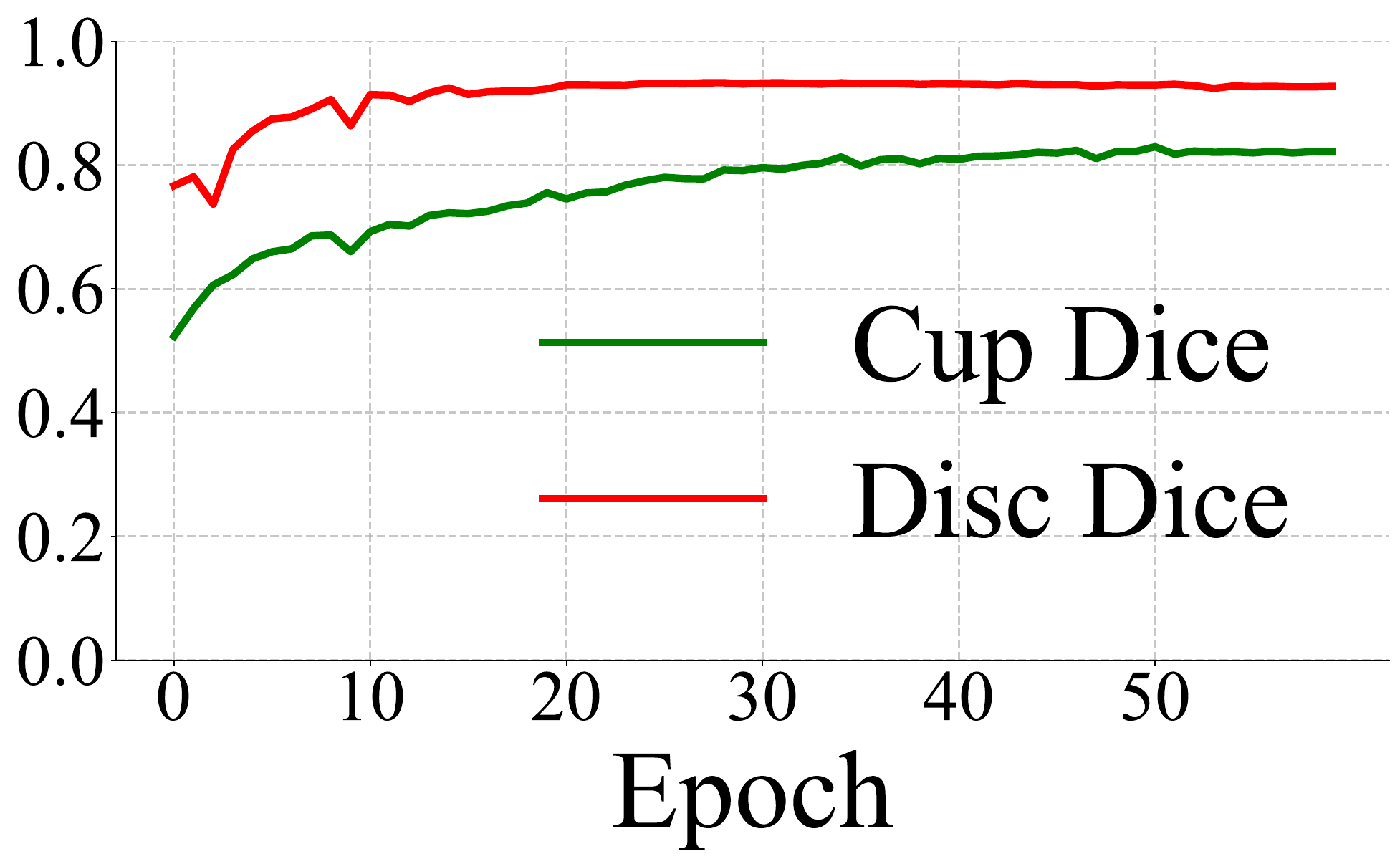}}
    \subfloat[Domain 2 Dice]{\includegraphics[width=0.5\columnwidth]{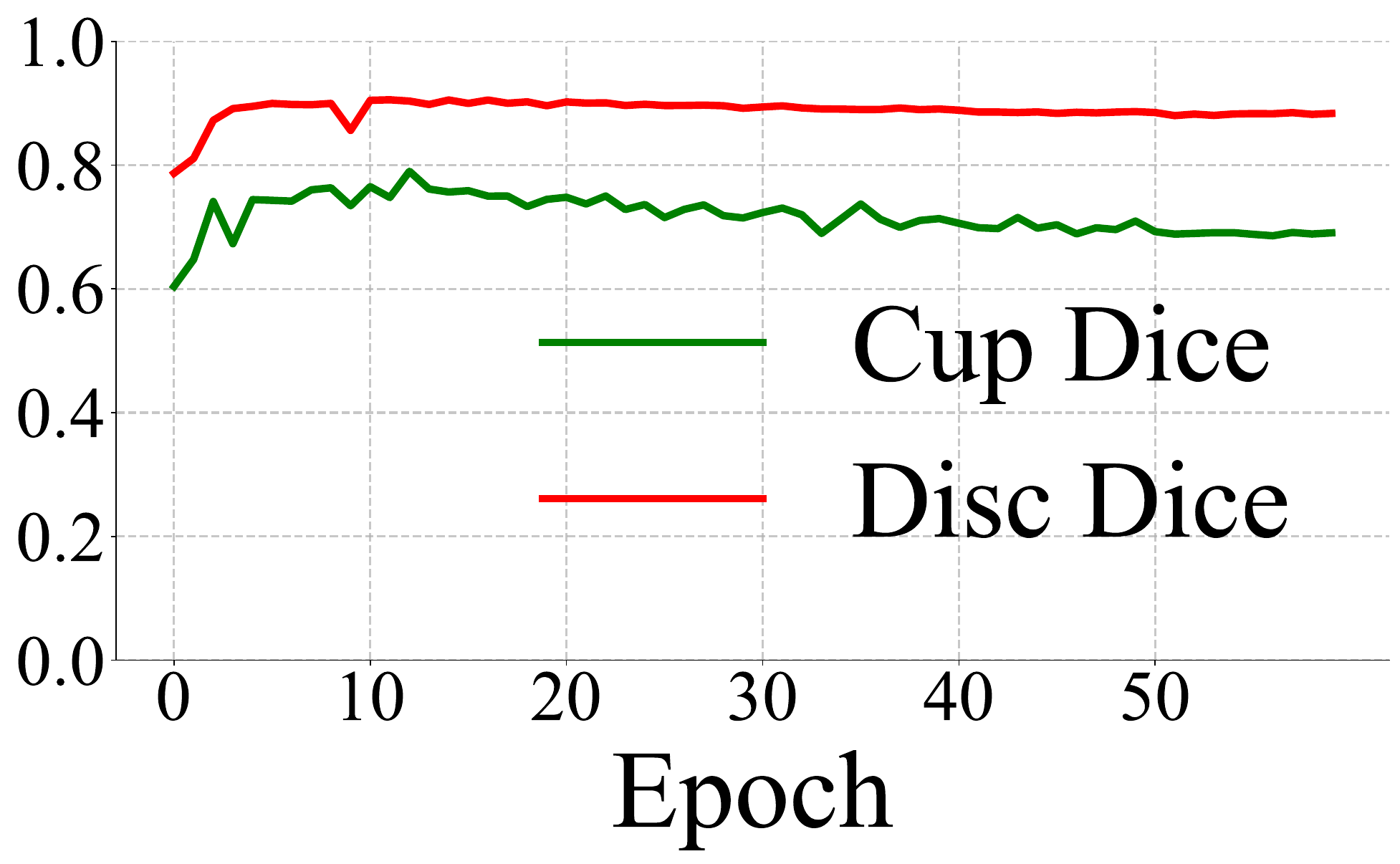}}
    \subfloat[Domain 3 Dice]{\includegraphics[width=0.5\columnwidth]{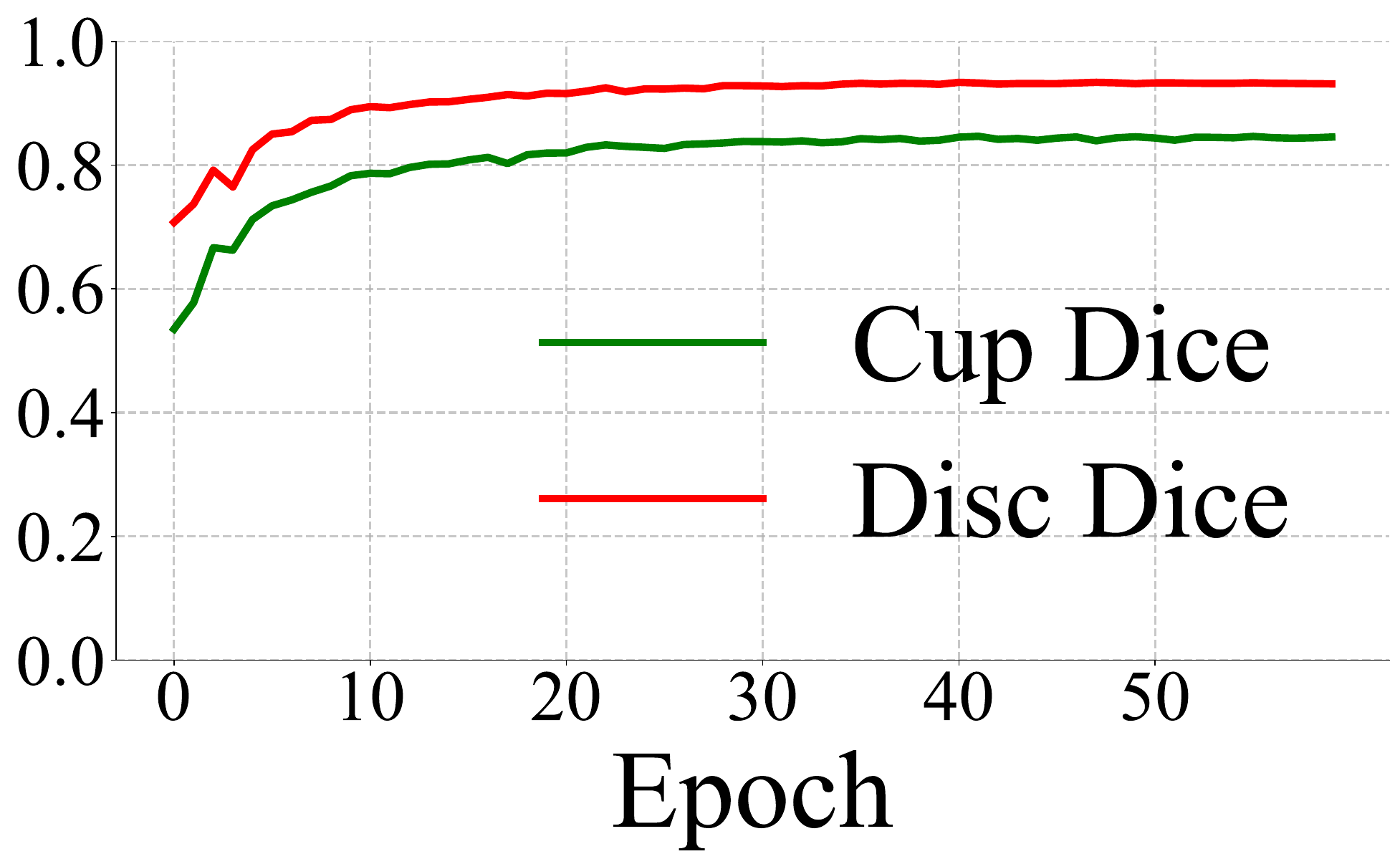}}
    \subfloat[Domain 4 Dice]{\includegraphics[width=0.5\columnwidth]{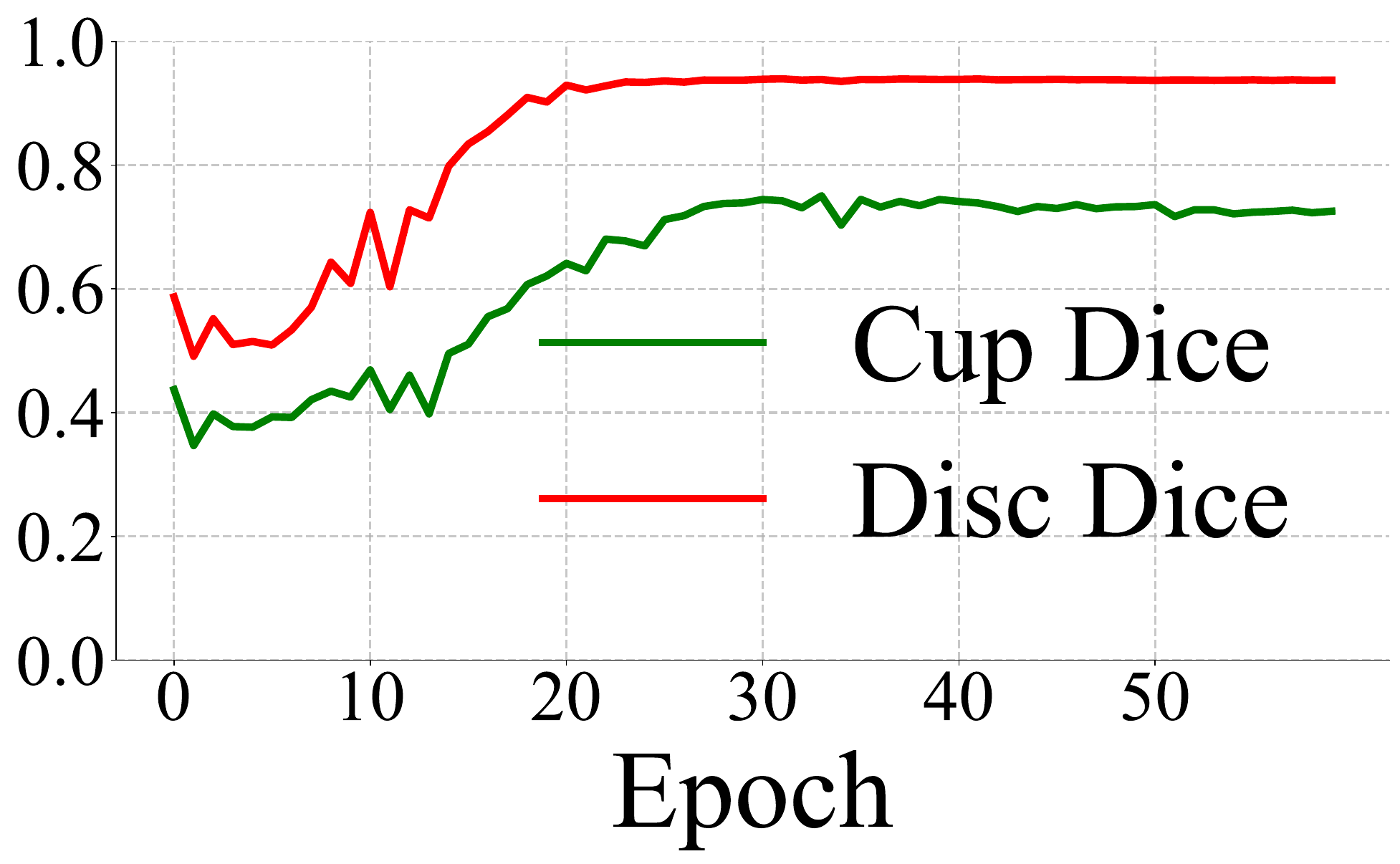}}
    \\
    
    \vspace{-2mm}
    \caption{Validation Loss and Dice across four domains on Fundus dataset.}
    \label{fig:loss_dice}
    \vspace{-2mm}
\end{figure*}

\begin{figure}[ht]
    \centering
    \subfloat[Domain 1 Loss]{\includegraphics[width=0.45\textwidth]{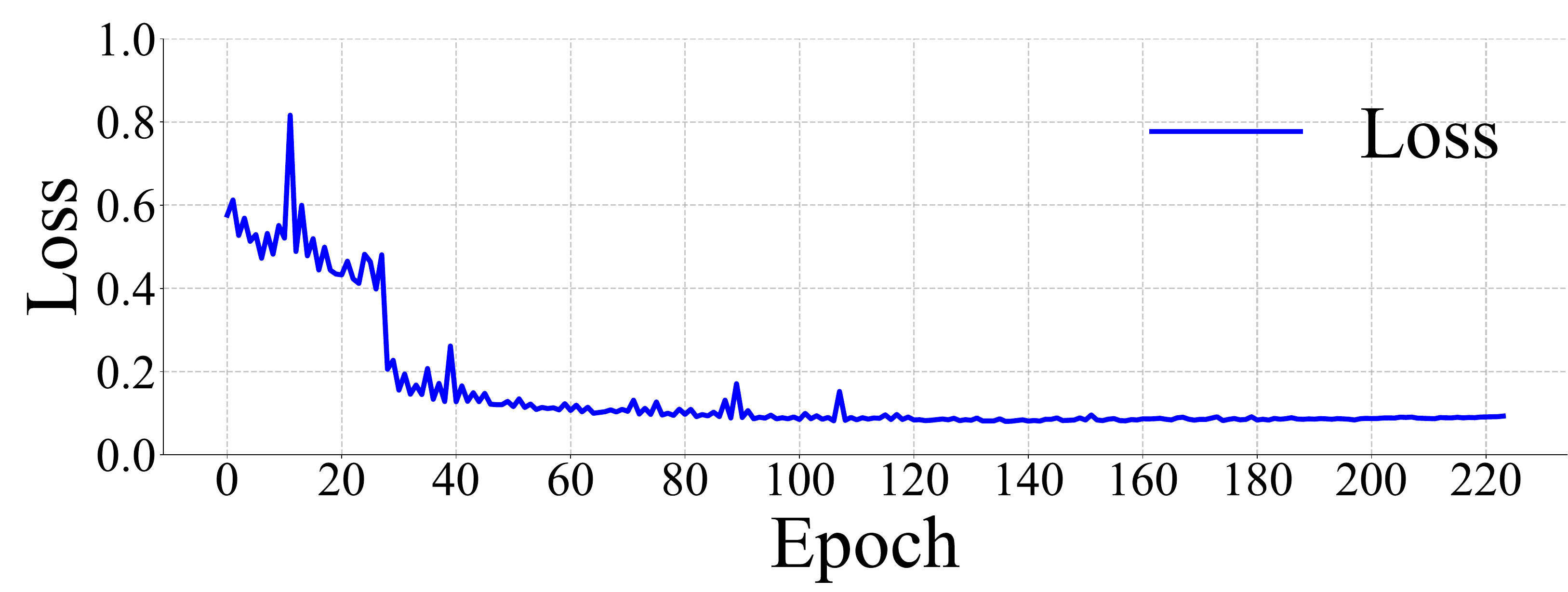}}
    \hspace{5mm} % 调整两张图之间的水平间距
    \subfloat[Domain 1 Dice]{\includegraphics[width=0.45\textwidth]{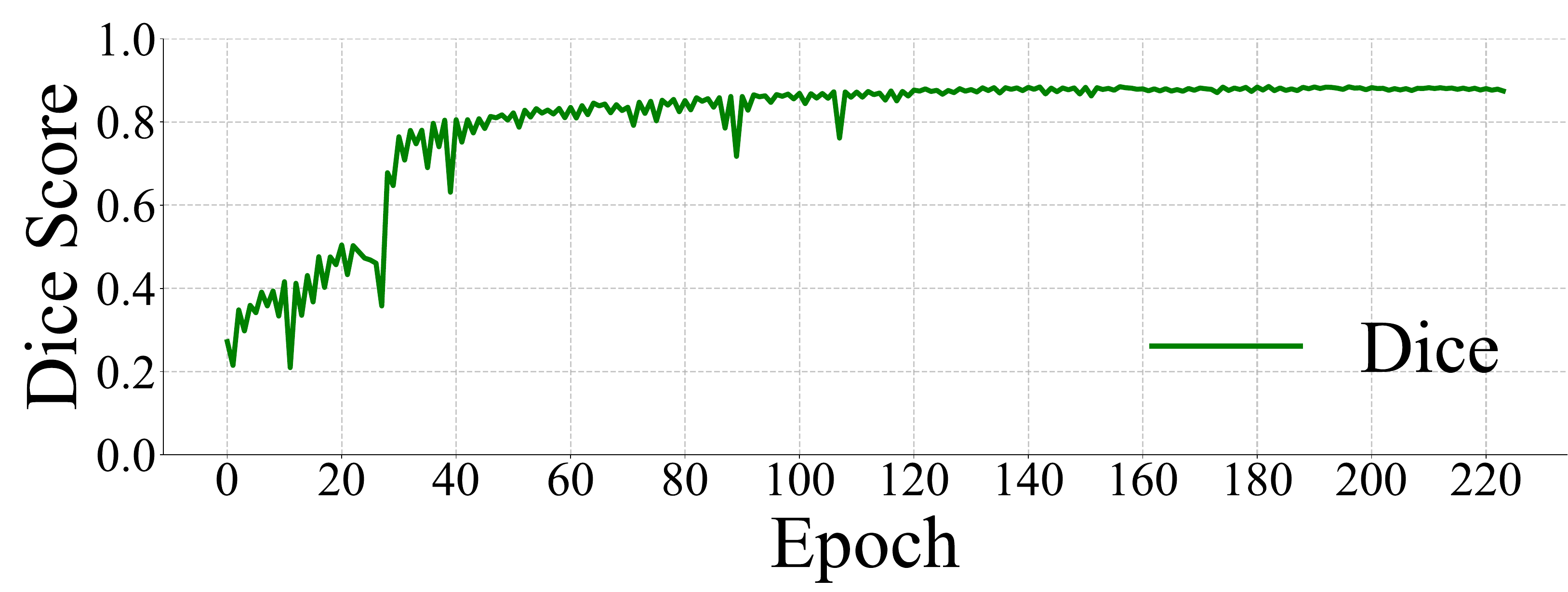}}
    \vspace{-2mm}
    \caption{Validation Loss and Dice for Domain 1 on the Prostate dataset.}
    \label{fig:loss_dice}
    \vspace{-2mm}
\end{figure}

\section{Conclusion}
This study addresses the challenge of data annotation in medical image segmentation by introducing a sharpness-aware optimization method based on $f$-divergence minimization (DiM) for semi-supervised learning. While existing semi-supervised methods (SSMIS), including sharpness-aware optimization (SAM), have shown success, they often overlook distribution differences between datasets. The proposed DiM method enhances model stability by adjusting the sensitivity of model parameters and improves adaptability to varying datasets. By reducing $f$-divergence, DiM achieves a better balance in performance between source and target datasets and mitigates overfitting. Experimental results demonstrate that DiM significantly improves performance, as evidenced by its groundbreaking progress in Dice scores on the prostate dataset, with similar success across three public datasets.
\clearpage
{
    \small
    \bibliographystyle{ieeenat_fullname}
    \bibliography{main}
}


\end{document}